\documentclass{llncs}
\usepackage{makeidx}  
\usepackage{amsmath}
\usepackage{amsfonts}
\usepackage{graphicx}
\usepackage{multirow}
\usepackage{algorithmic}
\usepackage{algorithm}
\usepackage{epstopdf}
\numberwithin{algorithm}{section}

\begin{document}

\mainmatter              
\title{HIPAD - A Hybrid Interior-Point Alternating Direction algorithm for knowledge-based SVM and feature selection}
\titlerunning{HIPAD}  
%
\author{Z. Qin\inst{1} \and X. Tang\inst{2} \and
I. Akrotirianakis\inst{3} \and A. Chakraborty\inst{3} }
\authorrunning{Qin et al.} 
%
\tocauthor{Z. Qin, X. Tang, I. Akrotirianakis, A. Chakraborty}
\institute{Columbia University, New York, NY,\\
\email{zq2107@columbia.edu}
\and
Lehigh University, Bethlehem, PA,\\
\email{xit210@lehigh.edu}
\and
Siemens Corporation, Corporate Technology, Princeton, NJ,\\
\email{(ioannis.akrotirianakis , amit.chakraborty)@siemens.com}
}

\maketitle              

\newtheorem{alg}{Algorithm}[section]
\newtheorem{thm}{Theorem}[section]
\newtheorem{cor}{Corollary}[section]
\newtheorem{lem}{Lemma}[section]
\newtheorem{rem}{Remark}[section]
\newcommand{\xbold}{\textbf{x}}
\newcommand{\Xbar}{\bar{X}}
\newcommand{\ybold}{\textbf{y}}
\newcommand{\xboldi}{\textbf{x}_i}
\newcommand{\xboldj}{\textbf{x}_j}
\newcommand{\xibar}{\overline{\xi}}
\newcommand{\alphabar}{\overline{\alpha}}
\newcommand{\alphatild}{\tilde{\alpha}}
\newcommand{\wbold}{\textbf{w}}
\newcommand{\cbold}{\textbf{c}}
\newcommand{\ubold}{\textbf{u}}
\newcommand{\vbold}{\textbf{v}}
\newcommand{\abold}{\textbf{a}}
\newcommand{\ebold}{\textbf{e}}
\newcommand{\zerobold}{\textbf{0}}
\newcommand{\dbold}{\textbf{d}}
\newcommand{\gbold}{\textbf{g}}
\newcommand{\zbold}{\textbf{z}}
\newcommand{\sbold}{\textbf{s}}
\newcommand{\tbold}{\textbf{t}}
\newcommand{\rbold}{\textbf{r}}

\begin{abstract}
We consider classification tasks in the regime of scarce labeled training data in high dimensional feature space, where specific expert knowledge is also available.  We propose a new hybrid optimization algorithm that solves the elastic-net support vector machine (SVM) through an alternating direction method of multipliers in the first phase, followed by an interior-point method for the classical SVM in the second phase.  Both SVM formulations are adapted to knowledge incorporation. Our proposed algorithm addresses the challenges of automatic feature selection, high optimization accuracy, and algorithmic flexibility for taking advantage of prior knowledge.  We demonstrate the effectiveness and efficiency of our algorithm and compare it with existing methods on a collection of synthetic and real-world data.

\end{abstract}
\section{Introduction}\label{sec:motivation}
Classification tasks on data sets with large feature dimensions are very common in real-world machine learning applications.  Typical examples include microarray data for gene selection and text documents for natural language processing.  Despite the large number of features present in the data sets, usually only small subsets of the features are relevant to the particular learning tasks, and local correlation among the features is often observed.  Hence, feature selection is required for good model interpretability.  Popular classification techniques, such as support vector machine (SVM) and logistic regression, are formulated as convex optimization problems. An extensive literature has been devoted to optimization algorithms that solve variants of these classification models with sparsity regularization \cite{pardalos-hansen-2008,sra-nowozin-wright-2011}.
Many of them are based on first-order (gradient-based) methods, mainly because the size of the optimization problem is very large. The advantage of first-order methods is that their computational and memory requirements at each iteration are low and as a result they can handle the large optimization problems occurring in classification problems. Their major disadvantage is their slow convergence, especially when a good approximation of the feature support has been identified. Second-order methods exhibit fast local convergence, but their computational and memory requirements are much more demanding, since they need to store and invert the Newton matrix at every iteration. It is therefore very important to be able to intelligently combine the advantages of both the first and the second order optimization methods in such a way that the resulting algorithm can solve large classification problems efficiently and accurately. As we will demonstrate in this paper such combination is possible by taking advantage of the problem structure and the change in its size during the solution process.  In addition, we will also show that our algorithmic framework is flexible enough to incorporate prior knowledge to improve classification performance.

\subsection{Related Work}\label{sec:related_work}
The above requirements demand three features from a learning algorithm: 1. it should be able to automatically select features which are possibly in groups and highly correlated;  2. it has to solve the optimization problem in the training phase efficiently and with high accuracy; and 3. the learning model needs to be flexible enough so that domain knowledge can be easily incorporated.  Existing methods are available in the literature that meet some of the above requirements {\em individually}.  For enforcing sparsity in the solution, efficient optimization algorithms such as that proposed in \cite{koh2007interior} can solve large-scale sparse logistic regression.  On the other hand, the $L_1$-regularization is unstable with the presence of highly correlated features - among a group of such features, essentially one of them is selected in a random manner.  To handle local correlation among groups of features, the elastic-net regularization \cite{zou2005regularization} has been successfully applied to SVM \cite{wang2006doubly} and logistic regression \cite{ryali2010sparse}.  However, incorporating domain knowledge into the logistic regression formulation is not straightforward.  For SVM, including such knowledge in the optimization process has been demonstrated in \cite{fung2003knowledge}. Recently, an Alternating Direction Method of Multipliers (ADMM) has been proposed for the elastic-net SVM (ENSVM) \cite{yeefficient}.  ADMM  is quick to find an approximate solution to the ENSVM problem, but it is known to converge very slowly to high accuracy optimal solutions  \cite{boyd2010distributed}.   The interior-point methods (IPM) for SVM are known to be able to achieve high accuracy in their solutions with a polynomial iteration complexity, and the dual SVM formulation is independent of the feature space dimensionality.  However, the classical $L_2$-norm SVM is not able to perform automatic feature selection.  Although the elastic-net SVM can be formulated as a QP (in the primal form), its problem size grows substantially with the feature dimensionality.  Due to the need to solve a Newton system in each iteration, the efficiency of IPM quickly deteriorates as the feature dimension becomes large.  

\subsection{Main Contributions}
In this paper we propose a new hybrid algorithmic framework for SVM to address {\em all} of the above challenges and requirements {\em simultaneously}.  Our framework combines the advantages of a first-order optimization algorithm (through the use of ADMM) and a second-order method (via IPM) to achieve both superior speed and accuracy.  Through a novel algorithmic approach that is able to incorporate expert knowledge, our proposed framework is able to exploit domain knowledge to improve feature selection, and hence, prediction accuracy.  Besides efficiency and generalization performance, we demonstrate through experiments on both synthetic and real data that our method is also more robust to inaccuracy in the supplied knowledge than existing approaches.

\section{A Two-phase Hybrid Optimization Algorithm}\label{sec:two_phase_alg}
As previously mentioned, for data sets with many features, the high dimensionality of the feature space still poses a computational challenge for IPM.  Fortunately, many data sets of this kind are very sparse, and the resulting classifier $w$ is also expected to be sparse, i.e. only a small subset of the features are expected to carry significant weights in classification.  Naturally, it is ideal for IPM to train a classifier on the most important features only.

Inspired by the Hybrid Iterative Shrinkage (HIS) \cite{shi2010fast} algorithm for training large-scale sparse logistic regression classifiers, we propose a two-phase algorithm to shrink the feature space appropriately so as to leverage the high accuracy of IPM while maintaining efficiency.  Specifically, we propose to solve an elastic-net SVM (ENSVM) or doubly-regularized SVM (DrSVM) \cite{wang2006doubly} problem during the first phase of the algorithm.  The elastic-net regularization performs feature selection with grouping effect and has been shown to be effective on data sets with many but sparse features and high local correlations \cite{zou2005regularization}.  This is the case for text classification, microarray gene expression, and fMRI data sets.  The support of the weight vector $w$ for ENSVM usually stabilizes well before the algorithm converges to the optimal solution.  Taking advantage of that prospect, we can terminate the first phase of the hybrid algorithm early and proceed to solve a classical SVM problem with the reduced feature set in the second phase, using an IPM solver.

\subsection{Solving the Elastic Net SVM using ADMM}\label{sec:admm}
SVM can be written in the regularized regression form as
\begin{equation}\label{eq:unconstr_svm}
    \min_{\wbold,b}\frac{1}{N}\sum_{i=1}^N(1 - (y_i(\xboldi^T\wbold + b)))_+ + \frac{\lambda}{2}\|\wbold\|_2^2,
\end{equation}
where the first term is an averaged sum of the hinge losses and the second term is viewed as a ridge regularization on $w$.  It is easy to see from this form that the classical SVM does not enforce sparsity in the solution, and $w$ is generally dense.  The ENSVM adds an $L_1$ regularization on top of the ridge regularization term, giving
\begin{equation}\label{eq:en_svm}
    \min_{\wbold,b}\frac{1}{N}\sum_{i=1}^{N}(1-y_i(\xboldi^T\wbold + b))_+ + \lambda_1\|\wbold\|_1 + \frac{\lambda_2}{2}\|\wbold\|_2^2.
\end{equation}
 Compared to the Lasso ($L_1$-regularized regression) \cite{tibshirani1996regression}, the elastic-net has the advantage of selecting highly correlated features in groups (i.e. the grouping effect) while still enforcing sparsity in the solution.  This is a particularly attractive feature for text document data, which is common in the hierarchical classification setting.  Adopting the elastic-net regularization as in (\ref{eq:en_svm}) brings the same benefit to SVM for training classifiers.

To approximately solve problem (\ref{eq:en_svm}), we adopt the alternating direction method of multipliers (ADMM) for elastic-net SVM recently proposed in \cite{yeefficient}.  ADMM has a long history dating back to the 1970s \cite{gabay1976dual}. Recently, it has been successfully applied to problems in machine learning \cite{boyd2010distributed}.  ADMM is a special case of the inexact augmented Lagrangian (IAL) method \cite{rockafellar1973multiplier} for the structured unconstrained problem
\begin{equation}\label{eq:struct_unc}
    \min_x F(x) \equiv f(x) + g(Ax),
\end{equation}
where both functions $f(\cdot)$ and $g(\cdot)$ are convex.  We can decouple the two functions by introducing an auxiliary variable $y$ and convert problem (\ref{eq:struct_unc}) into an equivalent constrained optimization problem
\begin{eqnarray}\label{eq:struct_c}
  \min_{x,y} && f(x) + g(y), \quad s.t.  \>\> Ax = y.
\end{eqnarray}
This technique is often called variable-splitting \cite{combettes2011proximal}.  The IAL method approximately minimizes in each iteration the augmented Lagrangian of (\ref{eq:struct_c}) defined by
    $\mathcal{L}(x,y,\gamma) := f(x) + g(y) + \gamma^T(y-Ax) + \frac{\mu}{2}\|Ax-y\|_2^2$
, followed by an update to the Lagrange multiplier $\gamma \gets \gamma - \mu(Ax - y)$.
The IAL method is guaranteed to converge to the optimal solution of (\ref{eq:struct_unc}), as long as the subproblem of approximately minimizing the augmented Lagrangian is solved with an increasing accuracy \cite{rockafellar1973multiplier}.  ADMM can be viewed as a practical implementation of IAL, where the subproblem is solved approximately by minimizing $\mathcal{L}(x,y;\gamma)$ with respect to $x$ and $y$ {\em alternatingly once}.
Eckstein and Bertsekas \cite{eckstein1992douglas} established the convergence of ADMM for the case of two-way splitting.
Now applying variable-splitting and ADMM to problem (\ref{eq:en_svm}), \cite{yeefficient} introduced auxiliary variables $(\abold,\cbold)$ and linear constraints so that the non-smooth hinge loss and $L_1$-norm in the objective function
are decoupled, making it easy to optimize over each of the variables.  Specifically, problem (\ref{eq:en_svm}) is transformed into an equivalent constrained form
\begin{eqnarray}\label{eq:en_svm_constr}
    \min_{\wbold,b,\abold,\cbold} && \frac{1}{N}\sum_{i=1}^N (a_i)_+ + \lambda_1\|\textbf{c}\|_1 + \frac{\lambda_2}{2}\|\wbold\|_2^2 \\
    \nonumber s.t. && \textbf{a} = \ebold - Y(X\wbold + b\ebold) \quad \mbox{and} \quad \cbold = \wbold
\end{eqnarray}
where $\xboldi^T$ is the $i$-th row of $X$, and $Y = \textrm{diag}(\ybold)$.
The augmented Lagrangian
\begin{multline}
    \mathcal{L}(\wbold,b,\abold,\cbold,\gamma_1,\gamma_2) := \frac{1}{N}\sum_{i=1}^N a_i + \lambda_1\|\textbf{c}\|_1 + \frac{\lambda_2}{2}\|\wbold\|_2^2  + \gamma_1^T(\ebold - Y(X\wbold + b\ebold)-\abold)\\
    +\gamma_2^T(\wbold-\cbold)  + \frac{\mu_1}{2}\|\ebold - Y(X\wbold + b\ebold)-\abold\|_2^2 + \frac{\mu_2}{2}\|\wbold-\cbold\|_2^2
\end{multline}
is then minimized with respect to $(\wbold,b), \abold,$ and $\cbold$ sequentially in each iteration, followed by an update to the Lagrange multipliers $\gamma_1$ and $\gamma_2$.  The original problem is thus decomposed into three subproblems consisting of computing the proximal operator of the hinge loss function (with respect to $\abold$), solving a special linear system (with respect to $(\wbold,b)$ ), and performing a soft-thresholding operation (with respect to $\cbold$), which can all be done in an efficient manner. Due to lack of space in the paper, we have included the detailed solution steps in the Appendix (see Algorithm \ref{alg:admm-ensvm} ADMM-ENSVM), where we define by $\mathcal{S}_\lambda(\cdot)$ the proximal operator associated with the hinge loss
\begin{eqnarray*}
  \mathcal{S}_\lambda(\omega) &=& \left\{
                                    \begin{array}{ll}
                                      \omega - \lambda, & \hbox{$\omega > \lambda$;} \\
                                      0, & \hbox{$0 \leq \omega \leq \lambda$;} \\
                                      \omega, & \hbox{$\omega < 0$.}
                                    \end{array}
                                  \right.
\end{eqnarray*}
and $\mathcal{T}_\lambda(\omega) = sgn(\omega)\max\{0,|\omega|-\lambda \}$ is the shrinkage operator. 

\subsection{SVM via Interior-Point Method}\label{sec:ipm}
Interior Point Methods enjoy fast convergence rates for a wide class of QP problems. Their theoretical polynomial convergence ($O(n\log \frac{1}{\epsilon})$) was first established by Mizuno \cite{mizuno1994}. In addition, Andersen {\em et al} \cite{Andersen1996implement} showed that the number of iterations needed by IPMs to converge is $O(\log n)$, which demonstrates that their computational effort increases in a slower rate than the size of the problem.

Both the primal and the dual SVM are QP problems. The primal formulation of SVM \cite{vapnik2000nature} is defined as
\begin{eqnarray*}
  \textrm{(SVM-P)} \qquad   \min_{\wbold,b,\xi,\sbold} && \frac{1}{2}\wbold^T\wbold + c\ebold^T\xi \\
  \qquad s.t. && y_i(\wbold^T\xboldi - b) + \xi_i - s_i = 1, i=1,\dots,N, \\
  && \sbold \geq 0, \xi \geq 0.
\end{eqnarray*}
whereas the dual SVM has the form
\begin{eqnarray*}
  \textrm{(SVM-D)} \quad \min_{\alpha} && \frac{1}{2}\alpha^TQ\alpha - \ebold^T\alpha \\
  s.t. && \ybold^T\alpha = 0, \quad \mbox{and} \quad 0 \leq \alpha_i \leq c, \quad i = 1,\cdots,N,
\end{eqnarray*}
where $Q_{ij} = y_iy_j\xboldi^T\xboldj = \bar{X}\bar{X}^T$.
By considering the KKT conditions of (SVM-P) and (SVM-D), the optimal solution is given by
$\wbold = \bar{X}^T\alpha = \sum_{i\in SV}\alpha_i y_i \xboldi$,
where $SV$ is the set of sample indices corresponding to the support vectors.  The optimal bias term $b$ can be computed from the complementary slackness condition $\alpha_i(y_i(\xboldi^T\wbold + b)-1 + \xi_i) = 0$.

Whether to solve (SVM-P) or (SVM-D) for a given data set depends on its dimensions as well as its sparsity.  Even if $X$ is a sparse matrix, $Q$ in (SVM-D) is still likely to be dense, whereas the Hessian matrix in (SVM-P) is the identity.  The primal problem (SVM-P), however, has a larger variable dimension and more constraints.  It is often argued that one should solve (SVM-P) when the number of features is smaller than the number of samples, whereas (SVM-D) should be solved when the number of features is less than that of the samples. Since in the second phase of Algorithm \ref{alg:hipad} we expect to have identified a small number of promising features, we have decided to solve (SVM-D) by using IPM.
Solving (SVM-D) is realized through the OOQP \cite{gertz2003object} software package that implements a primal-dual IPM for convex QP problems.

\subsection{The Two-phase Algorithm}
Let us keep in mind that the primary objective for the first phase is to appropriately reduce the feature space dimensionality without impacting the final prediction accuracy.  As we mentioned above, the suitability of ADMM for the first phase depends on whether the support of the feature vector converges quickly or not.  On  an illustrative dataset from \cite{yeefficient}, which has 50 samples with 300 features each, ADMM converged in 558 iterations. The output classifier $\wbold$ contained only 13 non-zero features, and the feature support converged in approximately 50 iteration (see Figure \ref{fig:fea_supp_plots} in the Appendix for illustrative plots showing the early convergence of ADMM). Although the remaining more than 500 iterations are needed by ADMM in order to satisfy the optimality criteria, they do not offer any additional information regarding the feature selection process. Hence, it is important to monitor the change in the signs and indices of the support and terminate the first phase promptly.  In our implementation, we adopt the criterion used in \cite{shi2010fast} and monitor the relative change in the iterates as a surrogate of the change in the support, i.e.
\begin{equation}\label{eq:hipad_transition}
    \frac{\|\wbold^{k+1}-\wbold^k\|}{\max(\|\wbold^k\|,1)} < \epsilon_{tol}.
\end{equation}
We have observed in our experiments that when the change over the iterates is small, the evolution of the support indices stabilizes too.

Upon starting the second phase, it is desirable for IPM to warm-start from the corresponding sub-vector of the solution returned by ADMM.  It should also be noted that we apply IPM during the second phase to solve the classical $L_2$-regularized SVM (\ref{eq:unconstr_svm}), instead of the ENSVM (\ref{eq:en_svm}) in the first phase.  There are two main reasons for this decision.  First, although ENSVM can be reformulated as a QP, the size of the problem is larger than the classical SVM due to the additional linear constraints introduced by the $L_1$-norm.  Second, since we have already identified (approximately) the feature support in the first phase of the algorithm, enforcing sparsity in the reduced feature space becomes less critical.  The two-phase algorithm is summarized in Algorithm \ref{alg:hipad}.

\begin{algorithm}
\caption{HIPAD (Hybrid Interior Point and Alternating Direction method)}
\begin{algorithmic}[1]\label{alg:hipad}
\STATE Given $\wbold^0, b^0, \abold^0, \cbold^0, \ubold^0,$ and $\vbold^0$.
\STATE \textbf{PHASE 1: ADMM for ENSVM}
\STATE $(\wbold^{\textrm{ADMM}},b^{\textrm{ADMM}}) \gets \textrm{ADMM-ENSVM}(\wbold^0, b^0, \abold^0, \cbold^0, \ubold^0,\vbold^0)$
\STATE \textbf{PHASE 2: IPM for SVM}
\STATE $\widetilde{\wbold} \gets $ non-zero components of $\wbold^{\textrm{ADMM}}$
\STATE $(\widetilde{X}, \widetilde{Y}) \gets $ sub-matrices of $(X, Y)$ corresponding to the support of $\wbold^{\textrm{ADMM}}$
\STATE $(\wbold,b) \gets \textrm{SVM-IPM}(\widetilde{X}, \widetilde{Y}, \widetilde{\wbold}$), through (SVM-D).
\RETURN $(\wbold,b)$
\end{algorithmic}
\end{algorithm}

\section{Domain Knowledge Incorporation}\label{sec:enk-svm}
Very often, we have prior domain knowledge for specific classification tasks.  Domain knowledge is most helpful when the training data does not form a comprehensive representation of the underlying unknown population, resulting in poor generalization performance of SVM on the unseen data from the same population.  This often arises in situations where labeled training samples are scarce, while there is an abundance of unlabeled data.

For high dimensional data, ENSVM performs feature selection along with training to produce a simpler model and to achieve better prediction accuracy.  However, the quality of the feature selection depends entirely on the training data.  In pathological cases, it is very likely that the feature support identified by ENSVM does not form a good representation of the population.  Hence, when domain knowledge about certain features is available, we should take it into consideration during the training phase and include the relevant features in the feature support should them be deemed important for classification.

In this section, we explore and propose a new approach to achieve this objective. We consider domain knowledge in the form of class-membership information associated with features.
We can incorporate such information (or enforce such rules) in SVM by adding equivalent linear constraints to the SVM QP problem (KSVM) \cite{fung2003knowledge,lauer2008incorporating}.  To be specific, we can model the above information with the linear implication
\begin{equation}\label{eq:linear_implication}
    B\xbold \leq \textbf{d} \quad \Rightarrow \quad \wbold^T\xbold + b \geq 1,
\end{equation}
where $B \in \mathbb{R}^{k_1\times m}$ and $\dbold \in \mathbb{R}^{k_1}$.
It is shown in \cite{fung2003knowledge} that by utilizing the non-homogeneous Farkas theorem of the alternative, 
(\ref{eq:linear_implication}) can be transformed into the following equivalent system of linear inequalities with a solution $\ubold$
\begin{equation}\label{eq:equiv_lin_constr}
    B^T\ubold + \wbold = \textbf{0}, \quad \textbf{d}^T\ubold - b + 1 \leq 0, \quad \textbf{u} \geq \textbf{0}.
\end{equation}
Similarly, for the linear implication for the negative class membership we have:
\begin{equation}\label{eq:lin_imp_neg}
    D\xbold \leq \gbold  \Rightarrow  \wbold^T\xbold + b \leq -1, \quad D \in \mathbb{R}^{k_2\times m}, g\in\mathbb{R}^{k_2},
\end{equation}
which can be represented by the set of linear constraints in $\vbold$
\begin{equation}\label{eq:equiv_lin_constr_neg}
    D^T\vbold - \wbold = \zerobold, \quad \gbold^T\vbold + b + 1 \leq 0, \quad \vbold \geq \zerobold.
\end{equation}
Hence, to incorporate the domain knowledge represented by (\ref{eq:linear_implication}) and (\ref{eq:lin_imp_neg}) into SVM, Fung {\em et al} \cite{fung2003knowledge} simply add the linear constraints (\ref{eq:equiv_lin_constr}) and (\ref{eq:equiv_lin_constr_neg}) to (SVM-P).
Their formulation, however, increases both the variable dimension and the number of linear constraints by at least $2m$, where $m$ is the number of features in the classification problem we want to solve.  This is clearly undesirable when $m$ is large, which is the scenario that we consider in this paper.

In order to avoid the above increase in the size of the optimization probelm, we choose to penalize the quadratics $\|B^T\ubold+\wbold\|_2^2$ and $\|D^T\vbold-\wbold\|_2^2$ instead of their $L_1$ counterparts considered in \cite{fung2003knowledge}. By doing so the resulting problem is still a convex QP but with a much smaller size.  Hence, we consider the following model for domain knowledge incorporation.
\begin{eqnarray*}
   \textrm{(KSVM-P)} \qquad
    \min_{\wbold,b,\xi,\ubold,\vbold,\eta_u,\eta_v} && \frac{1}{2}\wbold^T\wbold + c\ebold^T\xi + \frac{\rho_1}{2}\|B^T\ubold+\wbold\|_2^2 \\
    && + \rho_2\eta_u + \frac{\rho_3}{2}\|D^T\vbold-\wbold\|_2^2 + \rho_4\eta_v \\
  \nonumber s.t. && y_i(\wbold^T\xboldi + b) \geq 1 - \xi_i, \quad i = 1,\cdots,N, \\
   \nonumber && \textbf{d}^T\ubold - b + 1 \leq \eta_u, \\
   \nonumber && \textbf{g}^T\vbold + b + 1 \leq \eta_v,\\
   \nonumber && (\xi,\ubold,\vbold,\eta_u,\eta_v) \geq \zerobold.
\end{eqnarray*}

We are now ready to propose a novel combination of ENSVM and KSVM, and we will explain in the next section how the combined problem can be solved in our HIPAD framework.  The main motivation behind this combination is to exploit domain knowledge to improve the feature selection, and hence, the generalization performance of HIPAD.  To the best of our knowledge, this is the first method of this kind.

\subsection{ADMM Phase}
Our strategy for solving the elastic-net SVM with domain knowledge incorporation is still to apply the ADMM method.  First, we combine problems (\ref{eq:en_svm}) and (KSVM-P) and write the resulting optimization problem in an equivalent unconstrained form (by penalizing the violation of the inequality constraints through hinge losses in the objective function)
 \begin{equation*}
    \textrm{(ENK-SVM)} \quad \min_{\wbold,b,\ubold \geq \zerobold,\vbold \geq \zerobold} F(\wbold,b,\ubold,\vbold),
 \end{equation*}
where $F(\wbold,b,\ubold,\vbold) \equiv \frac{\lambda_2}{2}\|\wbold\|_2^2 + \lambda_1\|\wbold\|_1 + \frac{1}{N}\sum_{i=1}^{N}(1-y_i(x_i^Tw + b))_+
    + \frac{\rho_1}{2}\|B^T\ubold+\wbold\|_2^2 + \rho_2(\dbold^T\ubold - b + 1)_+ + \frac{\rho_3}{2}\|D^T\vbold-\wbold\|_2^2
    + \rho_4(\gbold^T\vbold + b + 1)_+$.
We then apply variable-splitting to decouple the $L_1$-norms and hinge losses and obtain the following equivalent constrained optimization problem:
\begin{eqnarray}\label{eq:ensvm_ksvm_c}
  \min_{\wbold,b,\ubold,\vbold,\abold,\cbold,p,q} &&\>\>\>  F(\wbold,b,\ubold,\vbold,\abold,\cbold,p,q) \\
  \nonumber s.t. && \>\>\> \abold = \ebold - (\bar{X}\wbold + \ybold b), \>\>\>  \cbold = \wbold, \\
  \nonumber && \>\>\> q = \dbold^T\ubold - b + 1,   \>\>\> p = \gbold^T\vbold + b + 1, \\
  \nonumber && \>\>\> \ubold \geq \zerobold, \>\>\> \vbold \geq \zerobold.
\end{eqnarray}
with $F(\wbold,b,\ubold,\vbold,\abold,\cbold,p,q)
  \equiv \frac{\lambda_2}{2}\|\wbold\|_2^2 + \lambda_1\|\cbold\|_1 + \frac{1}{N}\ebold^T(\abold)_+
    + \frac{\rho_1}{2}\|B^T\ubold+\wbold\|_2^2 + \rho_2(q)_+
+ \frac{\rho_3}{2}\|D^T\vbold-\wbold\|_2^2 + \rho_4(p)_+$.
As usual, we form the augmented Lagrangian $\mathcal{L}$ of problem (\ref{eq:ensvm_ksvm_c}),
\begin{multline*}
    \mathcal{L} := F(\wbold,b,\ubold,\vbold,\abold,\cbold,p,q)
    + \gamma_1^T(\ebold-(\Xbar\wbold+\ybold b)-\abold)
    + \frac{\mu_1}{2}\|\ebold-(\Xbar+\ybold b)-\abold\|_2^2\\
    + \gamma_2^T(\wbold-\cbold) + \frac{\mu_2}{2}\|\wbold-\cbold\|_2^2
    + \gamma_3(\dbold^T\ubold-b+1-q) + \frac{\mu_3}{2}\|\dbold^T\ubold-b+1-q\|_2^2 \\
    + \gamma_4(\gbold^T\vbold+b+1-p) + \frac{\mu_4}{2}\|\gbold^T\vbold+b+1-p\|_2^2
\end{multline*}
and minimize $\mathcal{L}$ with respect to $\wbold,b,\cbold,\abold,p,q,\ubold,\vbold$ individually and in order.  For the sake of readability, we do not penalize the non-negative constraints for $\ubold$ and $\vbold$ in the augmented Lagrangian.

Given $(\abold^k, \cbold^k, p^k, q^k)$, solving for $(\wbold,b)$ again involves solving a linear system
\begin{equation}\label{eq:ksvm_admm_linsys_wb}
        \left(
          \begin{array}{cc}
            \kappa_1 I + \mu_1 X^TX & \mu_1 X^T\ebold \\
            \mu_1\ebold^TX & \mu_1 N + \kappa_2 \\
          \end{array}
        \right)
         \left(
          \begin{array}{c}
            \wbold^{k+1} \\
            b^{k+1} \\
          \end{array}
        \right) =
        \left(
          \begin{array}{c}
             \textbf{r}_{\wbold}\\
             \textbf{r}_b\\
          \end{array}
        \right),
\end{equation}
where $\kappa_1 = \lambda_2 + \mu_2 + \rho_1 + \rho_3, \kappa_2 = \mu_3 + \mu_4, \textbf{r}_{\wbold} = X^TY\gamma_1^k + \mu_1 X^TY(\ebold-\abold^k) - \gamma_2^k + \mu_2\cbold^k + \rho_3D^T\vbold^k - \rho_1B^T\ubold^k$ and $\textbf{r}_{b} = \ebold^TY\gamma_1^k + \mu_1\ebold^TY(\ebold-\abold^k) + \gamma_3^k + \mu_3(\dbold^T\ubold^k + 1 - q^k) - \gamma_4^k - \mu_4(\gbold^T\vbold^k+1-p^k)$.
Similar to solving the linear system in Algorithm \ref{alg:admm-ensvm} ADMM-ENSVM, we can compute the solution to the above linear system through a few PCG iterations, taking advantage of the fact that the left-hand-side matrix is of low-rank.

To minimize the augmented Lagrangian with respect to $\ubold$, we need to solve a convex quadratic problem with non-negative constraints
\begin{equation}\label{eq:ksvm_admm_usub}
    \min_{\ubold \geq \zerobold} \quad \frac{\rho_1}{2}\|B^T\ubold + \wbold^{k+1}\|_2^2 + \gamma_3^k\dbold^T\ubold
    + \frac{\mu_3}{2}\|\dbold^T\ubold - b^{k+1} + 1 - q^{k}\|_2^2.
\end{equation}
Solving problem \eqref{eq:ksvm_admm_usub} efficiently is crucial for the efficiency of the overal algorithm. We describe a novel way to do so.
Introducing a slack variable $\sbold$ and transferring the non-negative constraint on $\ubold$ to $\sbold$, we decompose the problem into two parts which are easy to solve.  Specifically, we reformulate (\ref{eq:ksvm_admm_usub}) as
\begin{eqnarray*}
    \min_{\ubold,\sbold \geq \zerobold} && \>\>\> \frac{\rho_1}{2}\|B^T\ubold + \wbold^{k+1}\|_2^2 + \gamma_3^k\dbold^T\ubold
    + \frac{\mu_3}{2}\|\dbold^T\ubold - b^{k+1} + 1 - q^{k}\|_2^2 \\
    \nonumber s.t. && \>\>\>  \ubold - \sbold = \zerobold.
\end{eqnarray*}
Penalizing the linear constraint $\ubold - \sbold = \zerobold$ in the new augmented Lagrangian, the new subproblem with respect to $(\ubold,\sbold)$ is
\begin{multline}\label{eq:ksvm_admm_ussub}
    \min_{\ubold,\sbold \geq \zerobold} \quad \frac{\rho_1}{2}\|B^T\ubold + \wbold^{k+1}\|_2^2 + \gamma_3^k\dbold^T\ubold \\
    + \frac{\mu_3}{2}\|\dbold^T\ubold - b^{k+1} + 1 - q^k\|_2^2 + \gamma_5^T(\sbold-\ubold) + \frac{\mu_5}{2}\|\ubold-\sbold\|_2^2.
\end{multline}
Given an $\sbold^k \geq \zerobold$, we can compute $\ubold^{k+1}$ by solving a $k_1 \times k_1$ linear system
\begin{equation}\label{eq:ksvm_admm_linsys_u}
    (\rho_1 BB^T + \mu_3 \dbold\dbold^T + \mu_5 I)\ubold^{k+1} = \textbf{r}_\ubold,
\end{equation}
where $\textbf{r}_\ubold = -\rho_1B\wbold^{k+1} + \mu_3 \dbold b^{k+1} + \mu_3\dbold(q^k-1) - \dbold\gamma_3^k + \gamma_5 + \mu_5\sbold^k$.
We assume that $B$ has full row-rank.  This is a reasonable assumption since otherwise there is at least one redundant domain knowledge constraint and we can simply remove it. The number of domain knowledge constraints ($k_1$ and $k_2$) are usually small, so the system (\ref{eq:ksvm_admm_linsys_u}) can be solved exactly and efficiently by Cholesky factorization.

Solving for $\sbold^{k+1}$ corresponding to $\ubold^{k+1}$ is easy, observing that problem (\ref{eq:ksvm_admm_ussub}) is separable in the elements of $\sbold$.  For each element $s_i$, the optimal solution to the one-dimensional quadratic problem with a non-negative constraint on $s_i$ is given by $\max(0,u_i - \frac{(\gamma_5)_i}{\mu_5})$.  Writing in the vector form, $\sbold^{k+1} = \max(\zerobold, \ubold^{k+1} - \frac{\gamma_5^k}{\mu_5})$.
Similarly, we solve for $\vbold^{k+1}$ by introducing a non-negative slack variable $\tbold$ and solve the linear system
\begin{equation}\label{eq:ksvm_admm_linsys_v}
  (\rho_3DD^T + \mu_4 \gbold\gbold^T + \mu_6 I)\vbold^{k+1} = \rbold_\vbold,
\end{equation}
where $\rbold_\vbold = \rho_3D\wbold^{k+1} - \mu_4\gbold b^{k+1} - \gbold\gamma_4^k - \mu_4\gbold(1-p^k) + \gamma_6 + \mu_6\tbold^k$, and $\tbold^{k+1} = \max(0,\vbold^{k+1} - \frac{\gamma_6^k}{\mu_6})$.

Now given $(\wbold^{k+1},b^{k+1},\ubold^{k+1},\vbold^{k+1})$, the solutions for $\abold$ and $\cbold$ are exactly the same as in Lines \ref{line:a_sol} and \ref{line:c_sol} of Algorithm \ref{alg:hipad}, i.e.
\begin{eqnarray*}
  \abold^{k+1} &=& \mathcal{S}_{\frac{1}{N\mu_1}}\left( \ebold + \frac{\gamma_1^k}{\mu_1} - Y(X\wbold^{k+1} + b^{k+1}\ebold) \right), \\
  \cbold^{k+1} &=& \mathcal{T}_{\frac{\lambda_1}{\mu_2}}\left( \frac{\gamma_2^k}{\mu_2}+\wbold^{k+1} \right).
\end{eqnarray*}
The subproblem with respect to $q$ is
\begin{eqnarray}\label{eq:ksvm_admm_qsub}
    \nonumber \min_q && \quad \rho_2(q)_+ - \gamma_3^k q  + \frac{\mu_3}{2}\|\dbold^T\ubold^k - b^{k+1} + 1 - q\|_2^2 \equiv \\
     && \rho_2(q)_+
     + \frac{\mu_3}{2}\|q - (\dbold^T\ubold^k - b^{k+1} + 1 + \frac{\gamma_3^k}{\mu_3})\|_2^2.
\end{eqnarray}
The solution is given by a (one-dimensional) proximal operator associated with the hinge loss
\begin{equation}\label{eq:ksvm_admm_qsol}
    q^{k+1} = \mathcal{S}_{\frac{\rho_2}{\mu_3}}\left( \dbold^T\ubold^k - b^{k+1} + 1 + \frac{\gamma_3^k}{\mu_3} \right).
\end{equation}
Similarly, the subproblem with respect to $p$ is
\begin{equation*}
    \min_p \quad\quad \rho_4(p)_+ - \gamma_4^k p + \frac{\mu_4}{2}\|\gbold^T\vbold^k + b^{k+1} + 1 - p\|_2^2,
\end{equation*}
and the solution is given by
\begin{equation}\label{eq:ksvm_admm_psol}
    p^{k+1} = \mathcal{S}_{\frac{\rho_4}{\mu_4}}\left( \gbold^T\vbold^k + b^{k+1} + 1 + \frac{\gamma_4^k}{\mu_4} \right).
\end{equation}
 Due to lack of space in the paper, we summarize the detailed solution steps in the Appendix (see Algorithm \ref{alg:admm_enk} ADMM-ENK)

Although there appears to be ten additional parameters (six $\rho$'s and four $\mu$'s) in the ADMM method for ENK-SVM, we can usually set the $\rho$'s to the same value and do the same for the $\mu$'s.  Hence, in practice, there is only one additional parameter to tune, and our computational experience in Section \ref{sec:exp_ksvm} is that the algorithm is fairly insensitive to the $\mu$'s and $\rho$'s.

\subsection{IPM Phase}
The second phase for solving the knowledge-based SVM problem defined by (KSVM-P) follows the same steps as that described in section \ref{sec:ipm}. Note that in the knowledge-based case we have decided to solve the primal problem. This decision was based on extensive numerical experiments with both the primal and dual formulation which revealed that the primal formulation is more efficient.

We found in our experiments that by introducing slack variables and transforming the above problem into a linearly equality-constrained QP, Phase 2 of HIPAD usually requires less time to solve.

\subsection{HIPAD with domain knowledge incorporation}
We formally state the new two-phase algorithm for the elastic-net KSVM in Algorithm \ref{alg:hipad_ksvm}.
\begin{algorithm}
\caption{HIPAD-ENK}
\begin{algorithmic}[1]\label{alg:hipad_ksvm}
\STATE Given $\wbold^0, b^0, \abold^0, \cbold^0, \ubold^0,\vbold^0,p^0,q^0,\sbold^0\geq\zerobold,\tbold^0\geq\zerobold$.
\STATE \textbf{PHASE 1: ADMM for ENK-SVM}
\STATE $(\wbold,b,\ubold,\vbold) \gets $ADMM-ENK$(\wbold^0, b^0, \abold^0, \cbold^0, \ubold^0,\vbold^0,p^0,q^0,\sbold^0,\tbold^0)$
\STATE \textbf{PHASE 2: IPM for KSVM}
\STATE $\widetilde{\wbold} \gets $ non-zero components of $\wbold$
\STATE $(\widetilde{X}, \widetilde{Y}) \gets $ sub-matrices of $(X, Y)$ corresponding to the support of $\wbold$
\STATE $\eta_u^0 \gets \dbold^T\ubold-b+1$
\STATE $\eta_v^0 \gets \gbold^T\vbold+b+1$
\STATE $(\wbold,b) \gets \textrm{SVM-IPM}$$(\widetilde{X}, \widetilde{Y}, \widetilde{\wbold},b,\ubold,\vbold,\eta_u^0,\eta_v^0$)
\RETURN $(\wbold,b)$
\end{algorithmic}
\end{algorithm}

\section{Numerical results}
We present our numerical experience with the two main algorithms proposed in this paper: HIPAD and its knowledge-based version HIPAD-ENK. We compare their performance with their non-hybrid counterparts, i.e., ADMM-ENSVM and ADMM-ENK, which use ADMM to solve the original SVM problem.
The transition condition at the end of Phase 1 is specified in (\ref{eq:hipad_transition}), with $\epsilon_{tol} = 10^{-2}$.  The stopping criteria for ADMM are as follows: $\frac{|F^{k+1}-F^k|}{\max\{1,|F^k|\}} \leq \epsilon_1, \|\abold-(\ebold-\bar{X}\wbold-\ybold b)\|_2\leq \epsilon_1, \|\cbold-\wbold\|_2\leq \epsilon_1$ and $ \frac{\|\wbold^{k+1}-\wbold\|_2}{\|\wbold^k\|_2} \leq \epsilon_2$, with $\epsilon_1 = 10^{-5}$, and $\epsilon_2 = 10^{-3}$.

\subsection{HIPAD vs ADMM}
To demonstrate the practical effectiveness of HIPAD, we tested the algorithm on nine real data sets which are publicly available.  \textbf{rcv1} \cite{lewis2004rcv1} is a collection of manually categorized news wires from Reuters.  The original multiple labels have been consolidated into two classes for binary classification.  \textbf{real-sim} contains UseNet articles from four discussion groups, for simulated auto racing, simulated aviation, real autos, and real aviation.  Both \textbf{rcv1} and \textbf{real-sim} have large feature dimensions but are highly sparse.  The rest of the seven data sets are all dense data. \textbf{rcv1} and \textbf{real-sim} are subsets of the original data sets, where we randomly sampled 500 training instances and 1,000 test instances.  \textbf{gisette} is a handwriting digit recognition problem from NIPS 2003 Feature Selection Challenge, and we also sub-sampled 500 instances for training.  (For testing, we used the original test set of 1,000 instances.)  \textbf{duke}, \textbf{leukemia}, and \textbf{colon-cancer} are data sets of gene expression profiles for breast cancer, leukemia, and colon cancer respectively.  \textbf{fMRIa}, \textbf{fMRIb}, and \textbf{fMRIc} are functional MRI (fMRI) data of brain activities when the subjects are presented with pictures and text paragraphs.  The data was compiled and made available by Tom Mitchell's neuroinformatics research group \footnote{http://www.cs.cmu.edu/~tom/fmri.html}.  Except the three fMRI data sets, all the other data sets and their references are available at the LIBSVM website \footnote{http://www.csie.ntu.edu.tw/~cjlin/libsvmtools/datasets}.

The parameters of HIPAD, ADMM-ENSVM, and LIBSVM were selected through cross validation on the training data.  
We summarize the experiment results in Table \ref{tab:real_data_results}.
Clearly, HIPAD produced the best overall predication performance.  In order to test the significance of the difference, we used the test statistic in \cite{iman1980approximations} based on Friedman's $\chi^2_F$, and the results are significant at $\alpha=0.1$.  In terms of CPU-time, HIPAD consistently outperformed ADMM-ENSVM by several times on dense data.
The feature support sizes selected by HIPAD were also very competitive or even better than the ones selected by ADMM-ENSVM.  In most cases, HIPAD was able to shrink the feature space to below 10$\%$ of the original size.

\begin{table*}
\begin{center}
\resizebox{12.25cm}{!}{
\begin{tabular}{|c|c|c|c|c|c|c|c|}
  \hline
  \multirow{2}{*}{Data set} & \multicolumn{3}{|c|}{HIPAD} & \multicolumn{3}{|c|}{ADMM-ENSVM} & LIBSVM \\
  \cline{2-7}
  & Accuracy ($\%$) & Support size & CPU (s) & Accuracy ($\%$) & Support size & CPU (s) & accuracy ($\%$) \\
  \hline
  \textbf{rcv1} & \textbf{86.9} & 2,037 & 1.18 & 86.8 & 7,002 & 1.10 &  86.1 \\
  \textbf{real-sim} & \textbf{94.0} & 2,334 & 0.79 & 93.9 & 2,307 & 0.31 & 93.4 \\
  \textbf{gisette} & \textbf{94.7} & 498 & 8.96 & 63.1 & 493 & 45.87 & 93.4  \\
  \textbf{duke} & \textbf{90} & 168 & 1.56 & \textbf{90} & 150 & 5.52 & 80  \\
  \textbf{leukemia} & \textbf{85.3} & 393 & 1.70 & 82.4 & 717 & 6.35 & 82.4 \\
  \textbf{colon-cancer} & \textbf{84.4} & 195 & 0.45 & \textbf{84.4} & 195 & 1.34 & \textbf{84.4} \\
  \textbf{fMRIa} & \textbf{90} & 157 & 0.25 & \textbf{90} & 137 & 2.17 & 60 \\
  \textbf{fMRIb} & \textbf{90} & 45 & 0.23 & \textbf{90} & 680 & 0.75 & \textbf{90} \\
  \textbf{fMRIc} & \textbf{90} & 321 & 0.14 & \textbf{90} & 659 & 1.58 & \textbf{90} \\
  \hline
\end{tabular}
}
\end{center}
\caption{Experiment results of HIPAD and ADMM-ENSVM on real data.  The best prediction accuracy for each data set is highlighted in bold.}
\label{tab:real_data_results}
\end{table*}

\subsection{Simulation for Knowledge Incorporation}\label{sec:exp_ksvm}
We generated synthetic data to simulate the example presented at the beginning of Section \ref{sec:enk-svm} in the high dimensional feature space.  Specifically, four groups of multi-variate Gaussians $K_1,\cdots,K_4$ were sampled from $\mathcal{N}(\mu_1^+, \Sigma_1), \cdots, \mathcal{N}(\mu_4^+, \Sigma_4)$ and $\mathcal{N}(\mu_1^-, \Sigma_1), \cdots, \mathcal{N}(\mu_4^-, \Sigma_4)$ for four disjoint blocks of feature values ($\xbold_{K_1}, \cdots, \xbold_{K_4}$).  For positive class samples, $\mu_1^+ = 2, \mu_2^+ = 0.5, \mu_3^+ = -0.2, \mu_4^+ = -1$; for negative class samples, $\mu_1^- = -2, \mu_2^- = -0.5, \mu_3^- = 0.2, \mu_4^- = 1$.  All the covariance matrices have 1 on the diagonal and 0.8 everywhere else.  The training samples contain blocks $K_2$ and $K_3$, while all four blocks are present in the test samples.  A random fraction (5\%-10\%) of the remaining entries in all the samples are generated from the standard Gaussian distribution. 

The training samples are apparently hard to separate because the values of blocks $K_2$ and $K_3$ for the two classes are close to each other.  However, blocks $K_1$ and $K_4$ in the test samples are well-separated.  Hence, if we are given information about these two blocks as general knowledge for the entire population, we could expect the resulting SVM classifier to perform much better on the test data.  Since we know the mean values of the distributions from which the entries in $K_1$ and $K_4$ are generated, we can supply the following information about the relationship between the block sample means and class membership to the KSVM:
  $\frac{1}{L_1}\sum_{i\in K_1}x_i \geq 4 \Rightarrow \xbold \in A^+$, and
  $\frac{1}{L_4}\sum_{i\in K_4}x_i \geq 3 \Rightarrow \xbold \in A^-$
where $L_j$ is the length of the $K_j$, $j=1,\cdots,4$, $A^+$ and $A^-$ represent the positive and negative classes, and the lowercase $x_i$ denotes the $i$-th entry of the sample $\xbold$.  Translating into the notation of (KSVM-P), we have
\begin{equation*}
\small
  B = \left(
        \begin{array}{cccccc}
          \zerobold & \underbrace{-\frac{\ebold}{L_1}^T } & \zerobold & \zerobold & \zerobold &\zerobold \\
          & K_1 & & & & \\
        \end{array}
      \right)
   , d = -4, \label{eq:enk_data_knowledge1} \mbox{ and }
 D = \left(
        \begin{array}{cccccc}
          \zerobold & \zerobold & \zerobold & \zerobold & \underbrace{-\frac{\ebold}{L_4}^T }  & \zerobold \\
          & & & & K_4 & \\
        \end{array}
      \right)
   , g = -3.
\end{equation*}

The information given here is not precise, in that we are confident that a sample should belong to the positive (or negative) class only when the corresponding block sample mean well exceeds the distribution mean.  This is consistent with real-world situations, where the domain or expert knowledge tends to be conservative and often does not come in exact form.

We simulated two sets of synthetic data for ENK-SVM as described above, with $(N_{train} = 200, N_{test} = 400, m_{train} = 10,000)$ for \textbf{ksvm-s-10k} and $(N_{train} = 500, N_{test} = 1,000, m_{test} = 50,000)$ for \textbf{ksvm-s-50k}.  The number of features in each of the four blocks ($K_1,K_2,K_3,K_4$) is 50 for both data sets.
Clearly, HIPAD-ENK is very effective in terms of speed, feature selection, and prediction accuracy on these two data sets.  Even though the features in blocks $K_1$ and $K_4$ are not discriminating in the training data, HIPAD-ENK was still able to identify all the 200 features in the four blocks correctly and exactly.
This is precisely what we want to achieve as we explained at the beginning of Section \ref{sec:enk-svm}.  The expert knowledge not only helps rectify the separating hyperplane so that it generalizes better on the entire population, but also makes the training algorithm realize the significance of the features in blocks $K_1$ and $K_4$.
\begin{table*}
\begin{center}
\resizebox{12.25cm}{!}{
\begin{tabular}{|c|c|c|c|c|c|c|c|}
  \hline
  \multirow{2}{*}{Data set} & \multicolumn{3}{|c|}{HIPAD-ENK} & \multicolumn{3}{|c|}{ADMM-ENK} & LIBSVM \\
  \cline{2-7}
  & Accuracy ($\%$) & Support size & CPU (s) & Accuracy ($\%$) & Support size & CPU (s) & accuracy ($\%$) \\
  \hline
  \textbf{ksvm-s-10k} & \textbf{99} & 200 & 1.99 & 97.25 & 200 & 3.43 & 86.1 \\
  \textbf{ksvm-s-50k} & \textbf{98.8} & 200 & 8.37 & 96.4 & 198 & 20.89 & 74.8 \\
  \hline
\end{tabular}
}
\end{center}
\caption{Experiment results of HIPAD-ENK and ADMM-ENK on synthetic data.  The best prediction accuracy for each data set is highlighted in bold.}
\label{tab:enk_results}
\end{table*}

\section{Conclusion}
We have proposed a two-phase hybrid optimization framework for solving the ENSVM, in which the first phase is solved by ADMM, followed by IPM in the second phase.  In addition, we have proposed a knowledge-based extension of the ENSVM which can be solved by the same hybrid framework.  Through a set of experiments, we demonstrated that our method has significant advantage over the existing method in terms of computation time and the resulting prediction accuracy.  The algorithmic framework introduced in this paper is general enough and potentially applicable to other regularization-based classification or regression problems.

%
%
\bibliographystyle{abbrv}
\bibliography{scr_bib}

\newpage
\appendix
\section{Appendix}

\begin{algorithm}
\caption{ADMM-ENSVM}
\begin{algorithmic}[1]\label{alg:admm-ensvm}
\STATE Given $\wbold^0, b^0, \abold^0, \cbold^0, \gamma_1^0,$ and $\gamma_2^0$.
\FOR{$k = 0,1,\cdots,K-1$}
    \STATE $(\wbold^{k+1}, b^{k+1}) \gets$ PCG solution of:
        $\left(
          \begin{array}{cc}
            (\lambda_2 + \mu_2)I + \mu_1 X^TX & \mu_1 X^T\ebold \\
            \mu_1\ebold^TX & \mu_1 N \\
          \end{array}
        \right)
        \left(
          \begin{array}{c}
            \wbold^{k+1} \\
            b^{k+1} \\
          \end{array}
        \right) =
        \left(
          \begin{array}{c}
            X^TY\gamma_1^k - \mu_1 X^TY(\abold^k-\ebold) - \gamma_2^k + \mu_2\cbold^k \\
            \ebold^TY\gamma_1^k - \mu_1\ebold^TY(\abold^k - \ebold) \\
          \end{array}
        \right)$ \label{line:admm_linsys}

    \STATE $\abold^{k+1} \gets \mathcal{S}_{\frac{1}{N\mu_1}}\left( \ebold + \frac{\gamma_1^k}{\mu_1} - Y(X\wbold^{k+1} + b^{k+1}\ebold) \right)$ \label{line:a_sol}

    \STATE $\cbold^{k+1} \gets \mathcal{T}_{\frac{\lambda_1}{\mu_2}}\left( \frac{\gamma_2^k}{\mu_2}+\wbold^{k+1} \right)$ \label{line:c_sol}
    \STATE $\gamma_1^{k+1} \gets \gamma_1^k + \mu_1(\ebold - Y(X\wbold^{k+1} + b^{k+1}\ebold) - \abold^{k+1})$
    \STATE $\gamma_2^{k+1} \gets \gamma_2^k + \mu_2(\wbold^{k+1} - \cbold^{k+1})$
\ENDFOR
\RETURN $(\wbold^K,b^K)$
\end{algorithmic}
\end{algorithm}

\begin{figure}
\begin{center}
    \hspace*{-0.99in}\includegraphics[width=0.99\textwidth]{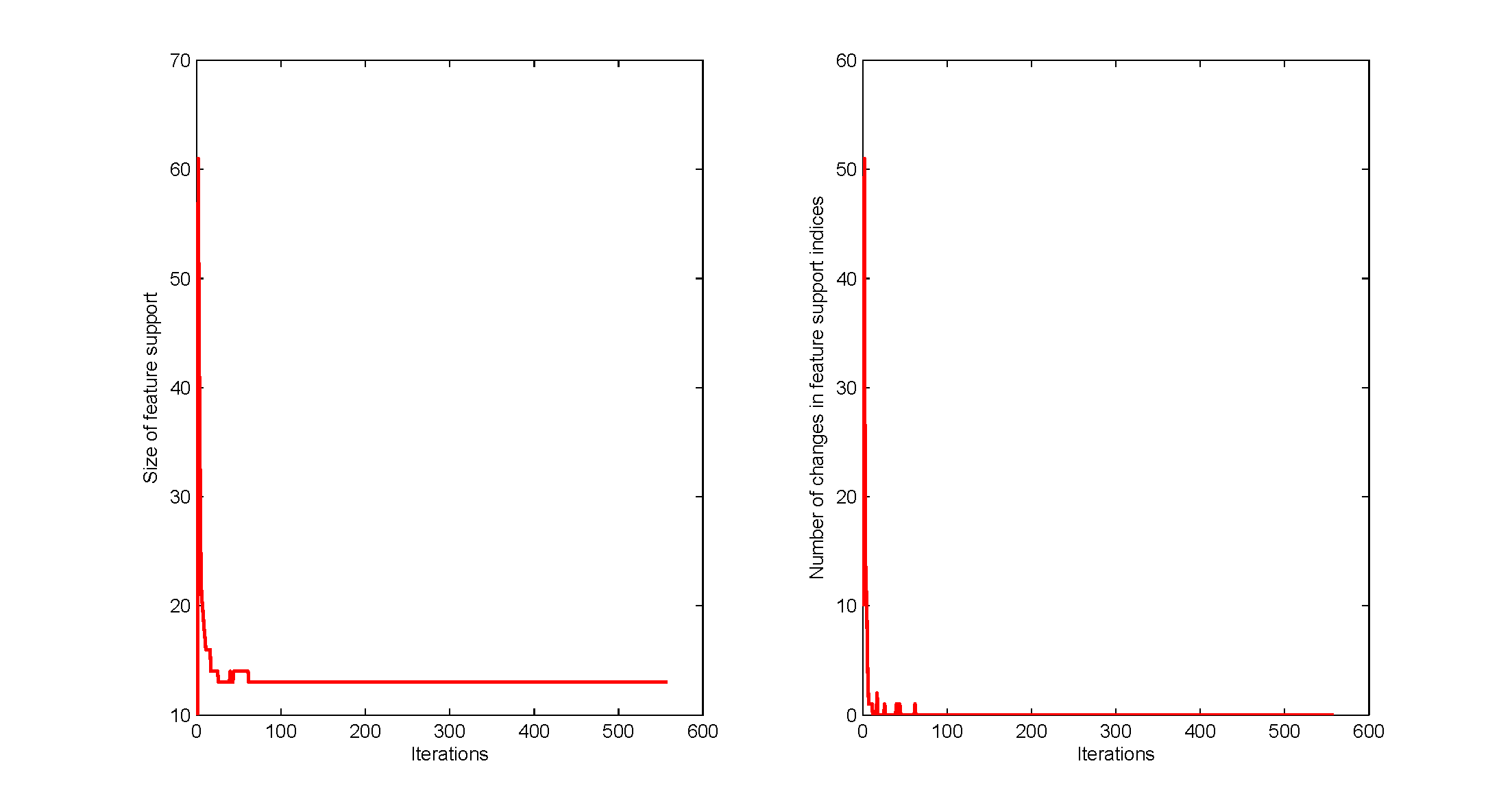}
    \caption{Illustration of the early convergence (in approximately 50 iterations) of the feature support for ADMM.}
    \label{fig:fea_supp_plots}
\end{center}
\end{figure}

\begin{algorithm}
\caption{ADMM-ENK}
\begin{algorithmic}[1]\label{alg:admm_enk}
\STATE Given $\wbold^0, b^0, \abold^0, \cbold^0, \ubold^0,\vbold^0,p^0,q^0,\sbold^0\geq\zerobold,\tbold^0\geq\zerobold,\gamma_i^0$, and the parameters $\lambda_1,\lambda_2,\rho_i, i = 1,\cdots,6$.
\FOR{$k = 0,1,\cdots,K-1$}
    \STATE $(\wbold^{k+1}, b^{k+1}) \gets $ PCG solution of the structured linear system \eqref{eq:ksvm_admm_linsys_wb}\label{line:admm_enk_linsys_wb}
    \STATE $\ubold^{k+1} \gets $ the solution of the linear system \eqref{eq:ksvm_admm_linsys_u}; $\>\>\>$ $\sbold^{k+1} \gets \max(\zerobold, \ubold^{k+1} - \frac{\gamma_5^k}{\mu_5})$
    \STATE $\vbold^{k+1} \gets $ the solution of the linear system \eqref{eq:ksvm_admm_linsys_v}; $\>\>\>$ $\tbold^{k+1} \gets \max(0,\vbold^{k+1} - \frac{\gamma_6^k}{\mu_6})$
    \STATE $\abold^{k+1} \gets \mathcal{S}_{\frac{1}{N\mu_1}}\left( \ebold + \frac{\gamma_1^k}{\mu_1} - Y(X\wbold^{k+1} + b^{k+1}\ebold) \right)$; $\>\>\>$ $\cbold^{k+1} \gets \mathcal{T}_{\frac{\lambda_1}{\mu_2}}\left( \frac{\gamma_2^k}{\mu_2}+\wbold^{k+1} \right)$
    \STATE $q^{k+1} \gets \mathcal{S}_{\frac{\rho_2}{\mu_3}}\left( \dbold^T\ubold^k - b^{k+1} + 1 + \frac{\gamma_3^k}{\mu_3} \right)$; $\>\>\>$ $p^{k+1} \gets \mathcal{S}_{\frac{\rho_4}{\mu_4}}\left( \gbold^T\vbold^k + b^{k+1} + 1 + \frac{\gamma_4^k}{\mu_4} \right)$
    \STATE $\gamma_1^{k+1} \gets \gamma_1^k + \mu_1(\ebold - Y(X\wbold^{k+1} + b^{k+1}\ebold) - \abold^{k+1})$; $\>\>\>$ $\gamma_2^{k+1} \gets \gamma_2^k + \mu_2(\wbold^{k+1} - \cbold^{k+1})$
    \STATE $\gamma_3^{k+1} \gets \gamma_3^k + \mu_3(\dbold^T\ubold^{k+1}-b^{k+1}+1-q^{k+1})$; $\>\>\>$ $\gamma_4^{k+1} \gets \gamma_4^k + \mu_4(\gbold^T\vbold^{k+1}+b^{k+1}+1-p^{k+1})$
    \STATE $\gamma_5^{k+1} \gets \gamma_5^k + \mu_5(\sbold^{k+1} - \ubold^{k+1})$; $\>\>\>$ $\gamma_6^{k+1} \gets \gamma_6^k + \mu_6(\tbold^{k+1} - \vbold^{k+1})$
\ENDFOR
\RETURN $(\wbold^K,b^K)$
\end{algorithmic}
\end{algorithm}

\end{document}